\documentclass[10pt,twocolumn,letterpaper]{article}

\usepackage[pagenumbers]{cvpr} 

\usepackage{graphicx}
\usepackage{amsmath}
\usepackage{amssymb}
\usepackage{booktabs}

\usepackage{xcolor}

\usepackage[pagebackref,breaklinks,colorlinks]{hyperref}

\usepackage[capitalize]{cleveref}
\crefname{section}{Sec.}{Secs.}
\Crefname{section}{Section}{Sections}
\Crefname{table}{Table}{Tables}
\crefname{table}{Tab.}{Tabs.}

\usepackage{multirow}

\def\cvprPaperID{****} 
\def\confName{CVPR}
\def\confYear{2023}

\begin{document}

\title{ActFormer: A GAN-based Transformer towards General Action-Conditioned 3D Human Motion Generation}
\author{
Liang Xu$^{2,3}$\footnotemark[1] \
Ziyang Song$^{4}$\footnotemark[1] \
Dongliang Wang$^{1,5}$ \
Jing Su$^1$\
Zhicheng Fang$^1$ \
Chenjing Ding$^1$ \\
Weihao Gan$^{1,5}$ \
Yichao Yan$^2$\ 
Xin Jin$^3$\
Xiaokang Yang$^2$\ 
Wenjun Zeng$^3$\
Wei Wu$^{1,5}$ \\
$^1$SenseTime Research \
$^2$Shanghai Jiao Tong University \
$^3$Eastern Institute for Advanced Study\\
$^4$The Hong Kong Polytechnic University \
$^5$Shanghai AI Laboratory}
\maketitle

\begin{abstract}

We present a GAN-based Transformer for general action-conditioned 3D human motion generation, including not only single-person actions but also multi-person interactive actions. 
Our approach consists of a powerful Action-conditioned motion TransFormer (ActFormer) under a GAN training scheme, equipped with a Gaussian Process latent prior.
Such a design combines the strong spatio-temporal representation capacity of Transformer, superiority in generative modeling of GAN, and inherent temporal correlations from the latent prior. 
Furthermore, ActFormer can be naturally extended to multi-person motions by alternately modeling temporal correlations and human interactions with Transformer encoders.
To further facilitate research on multi-person motion generation, we introduce a new synthetic dataset of complex multi-person combat behaviors.
Extensive experiments on NTU-13, NTU RGB+D 120, BABEL and the proposed combat dataset show that our method can adapt to various human motion representations and achieve superior performance over the state-of-the-art methods on both single-person and multi-person motion generation tasks, demonstrating a promising step towards a general human motion generator.

\renewcommand{\thefootnote}{\fnsymbol{footnote}} 
\footnotetext[1]{Denotes equal contribution. Work done when Liang and Ziyang was at SenseTime.}
\end{abstract}

\section{Introduction}
\label{sec:intro}

\begin{figure}[t]
  \centering
  \includegraphics[width=1.0\linewidth]{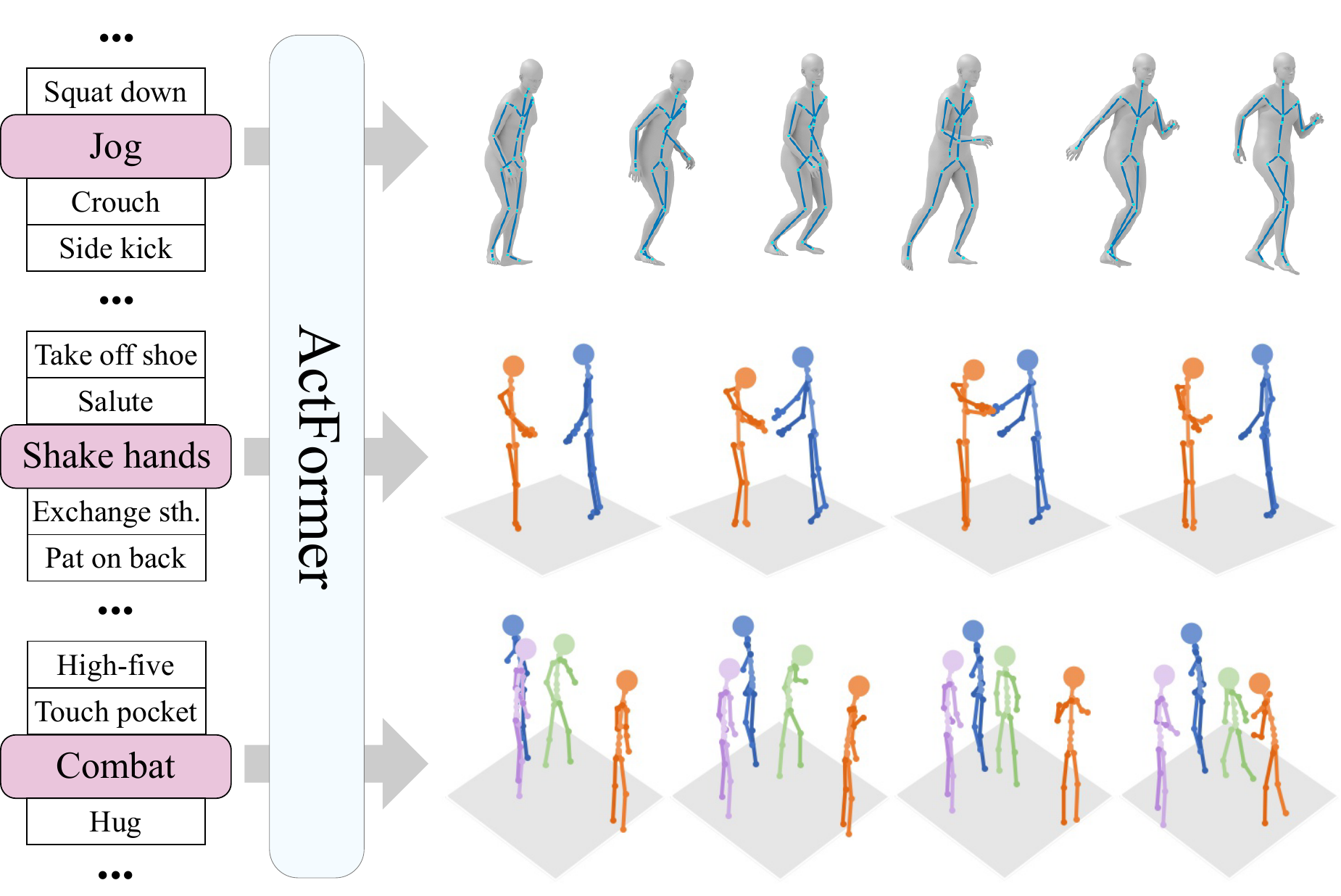}
  \caption{\textbf{Towards general action-conditioned 3D human motion generation.}
  Our framework adapts to more action categories, various human motion representations (\eg, SMPL body models, skeleton joint coordinates), and multi-person interactive actions.}
  \label{fig:intro}
  \vspace{-0.8cm}
\end{figure}

This work aims to tackle the action-conditioned motion generation task. Specifically, given a semantic action label as input and generate corresponding 3D human motions. The technique is key to applications like character animation creation, humanoid robots interaction and data synthesis for computer vision tasks related to human actions.

Human motion synthesis has been a long-standing research topic.
However, most of the prior works are closer to a prediction task, in which future motions are generated from previous motions~\cite{erd,structural_rnn,on_hmp,structured_pred,recurrent_vae,hp_gan,mt_vae,aist_pp}.
In recent years, some works started to focus on motion generation from action labels~\cite{action2motion,actor}. 
Despite some impressive generation results, these works are still limited in three aspects. 
Firstly, the generation is driven by small-scale human motion datasets, like NTU-13~\cite{nturgbd120,action2motion} and HumanAct12~\cite{action2motion}.
Each of those datasets covers only $\sim$10 action categories, thus the data distribution is relatively simple and easier to model.
There are larger datasets, including NTU RGB+D 120~\cite{nturgbd120} and BABEL~\cite{babel} with more complex data distribution, giving the chance to reinspect their capability.
Secondly, most of these works bias towards motion data of SMPL pose parameters while performing poorly on data of skeleton joint coordinates.
A solution adaptable to various human motion representations is thus expected.
Finally, prior works only focus on single-person motion generation while neglecting multi-person interactive actions, which is an integral part of daily human motions.
In general, prior works fail to cover a complete domain of human motions and stand far from a general human motion generator.

This paper explores a solution towards general action-conditioned human motion generation, as shown in~\cref{fig:intro}. 
The very first challenge lies in generating long motion sequences with realism and diversity.
Many prior works assume a Markovian dependency in temporal motions and adopt an auto-regressive model~\cite{perceptual_motion,action2motion,text2action,language2pose,dance2music}.
However, these methods are subject to the ``mean-pose'' problem, in which the model starts to generate the mean pose continuously after a few frames.
In contrast, CSGN~\cite{csgn} and ACTOR~\cite{actor} sample from a sequence-level latent prior and produce the whole sequence altogether.
Specifically, CSGN samples from a Gaussian Process (GP) latent prior and stacks convolutions in the generator to enforce temporal correlations.
On the other hand, ACTOR samples a single vector as the sequence-level embedding and produces multiple frames by querying through different positional encodings.
We argue that both are sub-optimal solutions, and we seek a better trade-off between inductive bias and representation capacity.
Our proposed \textbf{Act}ion-Conditioned Motion Trans\textbf{Former} (\textbf{ActFormer}) leverages the GP prior for the inherent temporal correlations.
Meanwhile, we adopt a Transformer architecture for its simple structure and strong power in encoding non-local correlations proved in many other tasks.
The Transformer model naturally regards a latent vector sequence from the GP prior as a sequence of tokens, leading to a seamless conjunction.
We incorporate the Transformer-based motion generator into GAN, known for high-quality generative modeling.
These designs jointly contribute to significant advantages of our framework in the single-person motion generation task.

Another challenge lies in handling human interactions when multi-person interactive actions are included.
Human interactions have been explored by some motion prediction algorithms, in which pooling or self-attention modules are adopted to encode the interactions~\cite{social_gan,social_context_pred,tripod,multi_agent_motion}.
However, it has not been considered in the motion generation task.
To our knowledge, our approach is the first to tackle multi-person motion generation.
We share the same latent vector sequence from GP among multiple persons in a group to enforce their synchronization over time.
Meanwhile, different persons are distinguished through positional encodings.
Our ActFormer can be easily extended to the multi-person scenario by alternately modeling temporal correlations and human interactions.
The generation results show impressive realism in both motions and multi-person interactions.

The strong demand for motion capture (MoCap) data with action labels also poses a challenge.
Prior methods rely on datasets with $\sim$10 categories, which can hardly drive a general motion generator.
MoCap datasets with multi-person interactive actions are even rarer. 
We leverage NTU RGB+D 120~\cite{nturgbd120} and the newly-released BABEL dataset~\cite{babel}, both including more than 100 action categories.
To facilitate the research on multi-person motion generation, we further construct a GTA Combat dataset through the Grand Theft Auto V's (GTA-V)~\cite{gta_v} gaming engine.
We collect $\sim$7K motion sequences of combat behavior, which is one of the most complex types of human interactions.
Experiments on these datasets verify the effectiveness of our approach.

Our three-fold contributions are summarized as follows:
(i) We propose ActFormer, a GAN-based Transformer framework, which adapts to various human motion representations and achieves leading results in the single-person motion generation task;
(ii) Our ActFormer takes a faithfully early step to solve the multi-person motion generation problem;
(iii) We contribute a GTA Combat dataset with plentiful and complex multi-person interactive motions.

\section{Related Work}

We review the literature on human motion prediction and generation tasks and MoCap datasets. 
We also review Transformers in GANs which are relevant to our approach.

\subsection{Motion Prediction}
Motion prediction aims to predict motions of future frames, given one or several frames of past motions.
Recurrent Neural Networks have been predominantly adopted to model sequence learning \cite{erd,structural_rnn,on_hmp,recurrent_vae,hp_gan,mt_vae,structured_pred}.
Especially, generative models like VAEs and GANs are incorporated in \cite{recurrent_vae,hp_gan,mt_vae}.
Recently, the powerful Transformer architecture is utilized by~\cite{aist_pp} to predict dance motions conditioned on music.

Multi-person interactions have also been considered in motion prediction.
\cite{social_gan} and \cite{social_context_pred} adopts the pooling module to aggregate information across multiple persons.
\cite{tripod} proposes a graph-based message passing mechanism to model both human-human and human-object interactions.
Recently, \cite{multi_agent_motion} uses the self-attention module to entangle multi-person motions.
Unlike the works above, we aim to tackle the motion generation task without relying on past motions.

\subsection{Motion Generation}
Compared to future motion prediction, motion generation from scratch is a relatively new and less explored field.
CSGN~\cite{csgn} generates unconstrained motions with a graph-convolution-based GAN framework.
\cite{perceptual_motion} further explores generating ever-changing motions for unbounded durations.
 
More works tend to generate motions from various conditions. \cite{dance2music,gen_dance_trans,dance_revolution} generates dance motions corresponding to the given music. \cite{lin:vigil18,ghosh2021synthesis,text2action,language2pose,teach,temos,flame} synthesize motions from language descriptions. More recently, diffusion model is also proposed for text-driven human motion generation in~\cite{motiondiffuse,motion_diffusion}.

Our work is dedicated to the task of action-conditioned motion generation, which uses semantic action labels as the condition.
Action2Motion~\cite{action2motion} and ACTOR~\cite{actor} are the works most similar to ours.
Action2Motion proposes a temporal VAE to generate motions frame by frame, based on GRU architecture.
ACTOR also adopts a VAE framework while leveraging the Transformer architecture and learning a sequence-level latent distribution, which differs from Action2Motion.
Our ActFormer also receives a sequence-level latent prior as input.
Unlike ACTOR, our input is a latent vector sequence sampled from a GP prior, inherently enforcing temporal correlations and thus reducing the difficulty of generating realistic motions. 

Another stream of works lies in simulation-based character control. 
For example, \cite{motion_vae} learns an autoregressive conditional VAE and then applies task-specific control policies based on this model.
 \cite{neural_state_machine} guides characters to achieve specific character-scene interaction goals with their motions.
Recently, \cite{scalable_control} and \cite{unicon} seek general approaches for physics-based character control.

\subsection{MoCap Dataset}
Large-scale MoCap data is critical for driving a general motion generator.
There have been many MoCap datasets~\cite{human36m,kit,mpi_hdm05,mpi_poselimits}, and AMASS~\cite{amass} contributes a large collection by unifying MoCap data from different sources into a common representation based on the SMPL body model~\cite{smpl}.
Despite a large amount of high-quality and diverse human motion data, semantic action labels are missing in AMASS.
Fortunately, the newly-released BABEL~\cite{babel} provides fine-grained action labels on frame-level for AMASS, resulting in a challenging and practically valuable benchmark for our task.

Some other datasets provide action labels, while their motion data is represented by skeleton joint coordinates.
Among them, NTU RGB+D 120~\cite{nturgbd120} is a large-scale and representative one.
Owing to the scalability of our method, we can learn from MoCap data with various motion representations.
The NTU RGB+D data is used for both single-person and multi-person motion generation.
As a complement to the daily interactive actions in NTU RGB+D, we contribute another GTA Combat dataset with more complex multi-person motions and interactions.

\subsection{Transformer in GANs}
The success of Transformers~\cite{transformer} in visual recognition tasks inspires its application in generation tasks.
TransGAN~\cite{transgan} is the first pure Transformer-based GAN architecture.
Later ViTGAN~\cite{vitgan} proposes novel regularization methods to improve the stability of Transformer-based GAN training.
 \cite{transg_convd} combines a Transformer-based generator and a CNN-based discriminator to form a robust model without cumbersome design choices.
We follow \cite{transg_convd} to exempt from the tricky designs in the discriminator and pay more attention to the generator.

\section{Approach}

\begin{figure*}[t]
  \centering
   \includegraphics[width=1.0\linewidth]{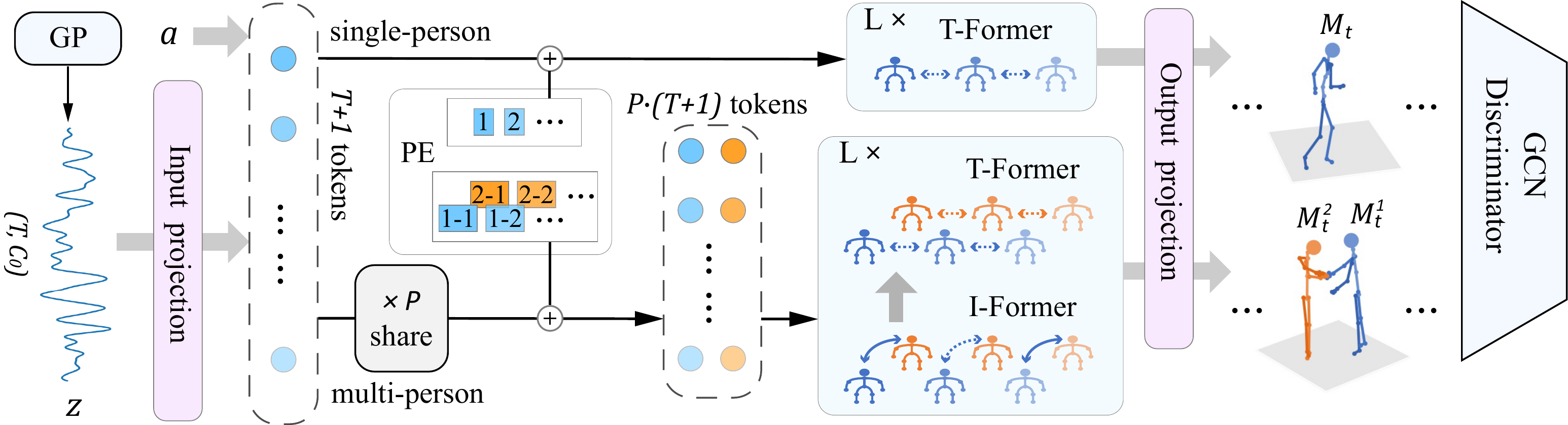}
   \caption{\textbf{Overview of the proposed ActFormer framework.}
   Given a latent vector sequence $z$ sampled from Gaussian Process (GP) prior and an action label $a$, the model can synthesize either a single-person (top stream) or a multi-person (bottom stream) motion sequence.
   The model is trained under a GAN scheme.}
   \label{fig:mogen}
   \vspace{-0.5cm}
\end{figure*}

The problem to be tackled is action-conditioned motion generation.
Formally speaking, given a semantic action label $a$ and a seed $z$ sampled from the latent prior, Our Action-Conditioned Motion TransFormer (ActFormer) will generate a sequence of human motions $M = \{ M_t | t \in \{ 1, ..., T \} \}$ corresponding to the action label.
Each frame $M_t$ contains the motions of $P$ persons, \ie, $M_t = \{ M^p_t | p \in \{ 1, ..., P \} \}$.
The motion of one person in a frame $M^p_t$ is composed of a root translation $l^p_t$ in a global coordinate frame, and local body poses ${\theta}^p_t$.
The latter can be either SMPL-based parameters or other formats like skeleton joint coordinates.
In this section, we first introduce our approach to solve the single-person motion generation and then show how it can be extended to the multi-person setting.

\subsection{Single-person Motion Generation}

The generation starts with sampling a random seed from the latent prior.
Since temporal correlation is critical to a realistic motion sequence, we select the Gaussian Process as our latent prior and sample a $(T, C_0)$ latent vector sequence $z$ for each generation.
The time length $T$ is the same as that of the motion sequence to be generated, and the latent vector at each time step has $C_0$ channels.
A $1$-d vector sequence with $T$ time steps is sampled independently from a GP on each of $C_0$ channels, with the characteristic length-scales on different channels spanning a spectrum of values.
As suggested by \cite{csgn}, this models a composition of correlations at various time scales into the latent vector sequence.

The ActFormer employs a Transformer-based generator to transform the latent vector sequence and the given action label, to a human motion sequence.
As shown in~\cref{fig:mogen}, at the input stage, the $(T, C_0)$ latent vector sequence $z$ is regarded as a list of $T$ tokens and passed through an MLP layer for input embedding.
To incorporate the semantic condition, we append a class token embedding the action category $a$, resulting in totally $T+1$ tokens.
Since the Transformer encoder is data-dependent, we add learnable positional encoding (PE) to the input tokens to maintain location information.
After that, $L$ layers of Temporal-transFormer (T-Former) model the temporal correlations among various time steps represented by tokens.
Finally, the class token is discarded while each of the rest $T$ tokens is projected by an output layer into a $C$-d vector.
The vector is regarded as a concatenation of a person's root translation and local body poses at a specific time step.

\subsection{Multi-person Motion Generation}

Transferring from the single-person to multi-person setting induces an additional \emph{person-wise} dimension $P$.
Fortunately, our approach introduced above can fit such an additional dimension $P$ by slight adjustment, thus be scaled to the multi-person case.

In the single-person setting, a $T$-frame motion sequence is encoded from a list of $T$ tokens.
Therefore, another $T$-frame motion sequence with $P$ persons requires $P$ lists of $T$ tokens.
Considering that motions of these $P$ persons are highly correlated and synchronized at each time step, we regard them as an entity and sample only one latent vector sequence to share among them.
Specifically, the $T$ input embedding tokens along with the class token are shared $P$ times to produce the input.
This strategy enforces the synchronization among multiple persons in a group from the input stage.
Meanwhile, different persons need to be distinguished from each other.
We turn to learnable positional encoding (PE) again to achieve this goal.
Here $T+1$ temporal positional encodings (TPE) and $P$ person-wise positional encodings (PPE) are separately learned.
Then they are tiled to generate $P \cdot (T+1)$ positional encodings to match input tokens.
Specifically, the PE for a $(t, p)$-indexed token $PE(t, p)=concat(TPE(t), PPE(p))$.
Completely independent PE for each token is also feasible, while such a 2D combination provides better results.

The ActFormer generator can be extended to a multi-person setting with slight adjustments, owing to the flexibility of Transformer encoders to adapt to modeling correlations along various dimensions.
As~\cref{fig:mogen} shows, an Interaction-transFormer (I-Former) firstly models the interactions among different persons at each time step independently.
Then a Temporal-transFormer (T-Former) follows to model the temporal correlations for each person independently.
Such a module, which alternately encodes human interactions and temporal correlations, is stacked $L$ layers.
Similar to the single-person case, all class tokens are discarded finally.
The remaining $P \cdot T$ tokens are projected into the final $(P, T, C)$ output, in which the $C$-d vector from each token represents the motion of a specific person at a particular time step.

\subsection{Generative Adversarial Training}

Our ActFormer is learned under the conditional generative adversarial training framework~\cite{gan,cgan}.
During training, the ActFormer generator synthesizes human motion sequences conditioned on given action labels.
Besides, a discriminator receives human motion sequences and action labels as inputs, trying to discriminate the generated human motions from real ones of specified actions.
The generator learns from the discriminator's feedback to improve its generation results to be close to real ones.
Conditional Wasserstein GAN loss functions~\cite{cgan,wgan} are adopted for training, formatted as,
\begin{eqnarray}
\begin{aligned}
    L_D &= \mathbb{E}[D(G(z, a), a) - D(\tilde{M}, a)], \\
    L_G &= \mathbb{E}[-D(G(z, a), a)], 
\end{aligned}
\end{eqnarray}
where $\tilde{M}$ represents the sequence sampled from real motion data belonging to the action category $a$.
$D$ denotes the discriminator and $G$ denotes the ActFormer generator.

An ST-GCN~\cite{stgcn} is adopted as the discriminator.
The $C$-d vector of a person's motion at each time step is decomposed into a $(K, D)$ part-wise representation.
The $K$-node Graph is constructed according to the skeleton topology behind the local body poses $\theta^p_t$, with the root translation $l^p_t$ also connected.
Therefore, a $(T, C)$ motion sequence is re-organized into a $(T, K, D)$ spatial-temporal graph.
In the multi-person setting, the part-wise motions of multiple persons are directly concatenated on the $D$ dimension.
In other words, the $(P, T, C)$ multi-person motion sequence becomes a $(T, K, P \cdot D)$ graph.
Finally, the graph-based motion sequence is input to the GCN discriminator to compute a score, with the semantic condition integrated by projection~\cite{cgan_projection}.

By concatenating the part-wise motions of multiple persons, the GCN can model their interactions at various spatial and temporal scales.
However, the concatenation operator is not permutation-invariant.
In other words, it cannot ensure to output the same score for the same motion sequence with various person-wise permutations.
We adopt a simple data augmentation strategy to compensate for this.
For each sample from the MoCap dataset, the persons inside are randomly permuted in every training iteration.
In this way, we encourage the ActFormer to regard the same sample with various permutations as different samples and model all of them into the learned distribution.
We find such a simple data augmentation more robust than importing symmetric functions into the discriminator network, since the latter often destabilizes the GAN training.

\section{Experiments}

In this section, we evaluate the proposed ActFormer on both single-person and multi-person motion generation tasks.
We firstly introduce the datasets and quantitative metrics used for evaluation.
Next, the ActFormer is compared with baseline methods from prior works.
Then we conduct an ablation study to investigate various components in ActFormer.
Finally, qualitative results are shown.

\subsection{Datasets and Evaluation Metrics}

\begin{figure}[t]
  \centering
  \includegraphics[width=1.0\linewidth]{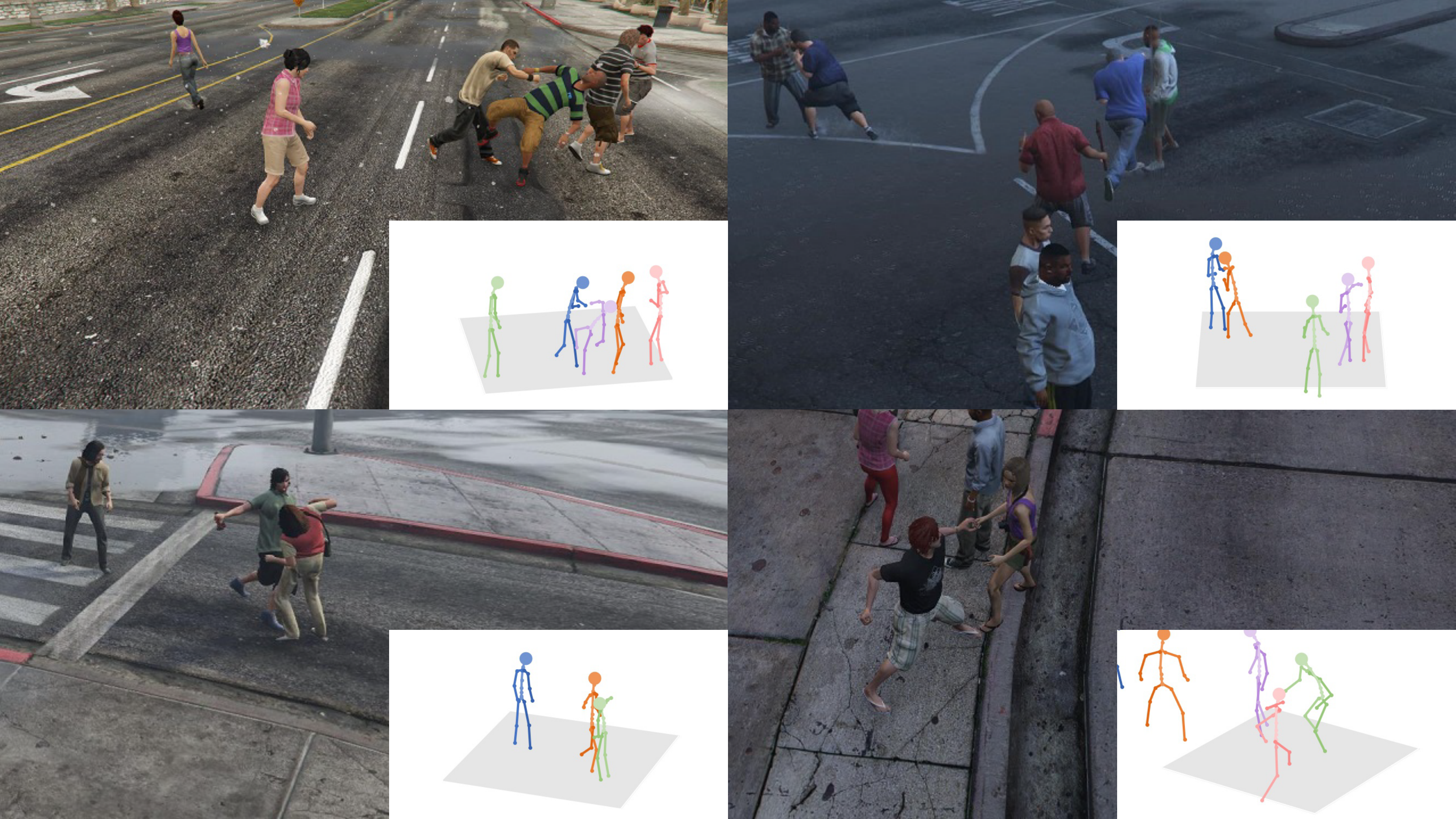}
  \caption{Sample RGB images and human pose annotations in the GTA Combat dataset.}
  \label{fig:gta_combat}
  \vspace{-0.5cm}
\end{figure}


\noindent{\textbf{NTU-13}~\cite{nturgbd120,action2motion}} is a subset of NTU RGB+D 120 with only 13 action categories and we choose it to be able to compare to previous works~\cite{action2motion,actor}. The SMPL-based motion data are obtained through VIBE~\cite{vibe}, as adopted in~\cite{action2motion,actor}.

\noindent{\textbf{NTU RGB+D 120}~\cite{nturgbd120}} contains 114,480 motion clips belonging to 120 action categories.
Among them, 94 categories are single-person actions, and the rest 26 categories are two-person interactive actions.
The dataset provides MoCap data in the format of skeleton joint coordinates, which are captured by Kinect~\cite{kinect} and severely noisy.
We fill missing detections and perform temporal smoothing to improve its quality.
Different parts of the dataset are used for both single-person and multi-person motion generation (referred to as NTU-1P and NTU-2P respectively in the following context).
Moreover, to evaluate the methods' adaptability to different motion representations, we apply a motion reconstruction pipeline similar to \cite{aist_pp} onto multi-view images from NTU-1P.
This results in a new version of NTU-1P data with SMPL parameters, named NTURecon-1P.

\noindent{\textbf{BABEL}~\cite{babel}} is another large-scale motion dataset with semantic action labels.
The SMPL-based motion data is derived from AMASS and has relatively high quality.
The whole dataset contains more than 250 action categories but is remarkably long-tailed.
Therefore, we follow the BABEL-120 benchmark in \cite{babel} and select motion data belonging to the most frequent 120 action categories.
The data is used for single-person motion generation.

\noindent{\textbf{GTA Combat}} is a synthetic multi-person MoCap dataset collected by us newly.
As there are few full-body MoCap datasets qualified for multi-person interactive action generation, we tackle the game GTA-V~\cite{gta_v} for synthetic data collection. 
We find combat behavior with 2 or more persons is a partial procedure generation process in GTA-V. 
The combat actions are randomly triggered by combing more than 10 atomic actions in real-time with more variations when equipped with different combating arms. 
And the attacked one can react randomly, triggered by the ragdoll physics~\cite{ragdoll} of the game engine. 
Overall, we can make actors in the game combat with each other and generate random samples with enough diversities. 
We sample 2 to 5 actors in each run and make each actor fight with one of the others randomly picked. 
We extend JTA~\cite{jta} which helps to extract 3D human skeletons with this random combat logic for the data collection. 
Thereby, we get a high-quality multi-person interaction MoCap dataset ranging from 2 to even 5 persons. 
\cref{fig:gta_combat} shows some samples of GTA Combat dataset.
More samples are provided in the supplementary.
According to the number of persons in each sequence, the dataset is divided into 4 splits.
Each split contains $\sim$2.3/1.9/1.5/1.2K sequences with motions of 2/3/4/5 persons.

\begin{table*}[t]
\footnotesize
  \centering
  \begin{tabular}{l c c c | c c c | c c c | c c c}
    \hline
    \noalign{\smallskip}
    & \multicolumn{3}{c}{NTU-13} & \multicolumn{3}{c}{NTURecon-1P} & \multicolumn{3}{c}{NTU-1P} &
    \multicolumn{3}{c}{BABEL} \\
    Method & Acc.$\uparrow$ & FID$_{m}$$\downarrow$ & FID$_{w}$$\downarrow$ &
    Acc.$\uparrow$ & FID$_{m}$$\downarrow$ & FID$_{w}$$\downarrow$ & Acc.$\uparrow$ & FID$_{m}$$\downarrow$ & FID$_{w}$$\downarrow$ & Acc.$\uparrow$ & FID$_{m}$$\downarrow$ & FID$_{w}$$\downarrow$ \\
    \noalign{\smallskip}
    \hline
    \noalign{\smallskip}
    Action2Motion~\cite{action2motion} & 94.9 & 4.40 & 2.01 & 41.97 & 23.86 & 18.35 & 8.34 & 56.13 & 46.13 & 6.04 & 17.03 & 6.05 \\
    CSGN~\cite{csgn} & 85.9 & 8.07 & 3.64 & 20.02 & 36.02 & 27.51 & 38.99 & 17.01 & 7.14 & 7.02 & 20.11 & 9.17 \\
    ACTOR~\cite{actor} & 97.1 & 5.35 & 1.18 & 39.69 & 37.87 & 19.58 & 35.56 & 32.89 & 9.63 & 3.44 & 55.66 & 36.27 \\
    Ours & \textbf{99.9} & \textbf{4.28} & \textbf{1.11} & \textbf{49.12} & \textbf{20.35} & \textbf{8.86} & \textbf{42.01} & \textbf{12.53} & \textbf{3.57} & \textbf{13.49} & \textbf{11.21} & \textbf{2.55} \\
    \noalign{\smallskip}
    \hline
  \end{tabular}
  \caption{\textbf{State-of-the-art comparison on single-person motion generation.}
  Higher Acc. and lower FIDs are better.}
  \label{tab:sota1p}
\end{table*}

\begin{table*}[t]
  \centering
  \begin{tabular}{l c c c c c}
    \hline
    \noalign{\smallskip}
    Method & Acc.$\uparrow$ & FID$_{m}$$\downarrow$ & FID$_{w}$$\downarrow$ & FID$^{a}_{m}$$\downarrow$ & FID$^{a}_{w}$$\downarrow$\\
    \noalign{\smallskip}
    \hline
    \noalign{\smallskip}
    Action2Motion~\cite{action2motion} & 16.85 & 15.07 & 10.12 & 26.25 & 20.31 \\
    CSGN~\cite{csgn} & 55.12 & 6.10 & 3.88 & 8.96 & 3.20 \\
    ACTOR~\cite{actor} & 63.04 & 13.56 & 6.72 & 19.14 & 8.16 \\
    Ours & \textbf{69.65} & \textbf{3.77} & \textbf{3.27} & \textbf{7.12} & \textbf{2.46} \\
    \noalign{\smallskip}
    \hline
  \end{tabular}
  \quad
  \begin{tabular}{l c c}
    \hline
    \noalign{\smallskip}
    Split & CSGN~\cite{csgn} & Ours \\
    \noalign{\smallskip}
    \hline
    \noalign{\smallskip}
    2p & 1.15 & \textbf{1.04} \\
    3p & 1.73 & \textbf{1.13} \\
    4p & 2.09 & \textbf{1.58} \\
    5p & 4.02 & \textbf{2.23} \\
    \noalign{\smallskip}
    \hline
  \end{tabular}
  \caption{\textbf{State-of-the-art comparison on multi-person motion generation.}
  \textbf{Left:} Performance on NTU-2P dataset.
  \textbf{Right:} FID$_a$ performance on different splits of the GTA Combat dataset.}
  \label{tab:sota2p}
  \vspace{-0.5cm}
\end{table*}

\noindent{\textbf{Evaluation Metrics.}}
We use action recognition accuracy and FID~\cite{fid} score as quantitative metrics.
Note that these metrics require a pre-trained action recognition model.
We adopt ST-GCN~\cite{stgcn}, sharing a similar configuration to the discriminator in the ActFormer's training.
Different from previous works~\cite{action2motion,actor}, the \textbf{root translation is considered} as an extra node when training the action recognition model since the root translation results are significant to measure the quality of the generated motions especially for the multi-person scenarios.

Previous works~\cite{action2motion,actor,csgn} consider statistics of the whole real/generated sample set when computing FID, regardless of action categories.
We denote this as whole FID (FID$_w$), while we find this not reasonable for the action-conditional generation case.
As a supplement, we adopt another mean FID (FID$_m$), in which FID is independently measured on samples of each action category and then averaged.
FID$_m$ assumes the conditional distributions of per-class samples to be individual Gaussians, thus better describing the similarity between two \emph{mixture} distributions in this task. We adopt FID$_m$ and FID$_w$ for all the experiments.

When extending to the multi-person setting, we measure the FID in two ways to observe generation results from different perspectives.
The first is to regard a group of persons as a whole.
Specifically, the recognition model receives concatenated multi-person motions as input and extracts features for calculating FID directly from it.
Data augmentation by random person-wise permutation is also applied in the recognition model's training to compensate for permutation invariance.
The second is to construct multi-person features by aggregating single-person features.
Towards this end, we train a single-person action recognition model.
During the evaluation, we extract features of each person in a group independently and aggregate them by channel-wise max pooling.
Such feature extraction is applied on both real and generated samples for alignment.
FID calculated in this way is called \emph{Aggregated FID} (FID$^{a}$).

\noindent{\textbf{Implementation Details.}}
For each dataset, we use different data splits to train ActFormer and action recognition models.
In NTU RGB+D 120, we follow its cross-subject split.
In BABEL, we also follow the provided split as~\cite{babel}.
The data distribution of different action categories in BABEL is remarkably long-tailed.
Therefore, we adopt a square-root sampling~\cite{square_root_sample} strategy in training the ActFormer, the action recognition models, and all the compared baselines.
GTA Combat contains only one action category, making it infeasible to train an action recognition model on it.
Also, to our knowledge, there are no public datasets whose motions contain 2$\sim$5 participants and meanwhile with action categories annotated.
Therefore, we leverage an action recognition model trained on NTU-1P by aligning motion data in NTU and GTA Combat into a unified skeleton topology.
In this way, FID$^{a}$ can be measured on GTA Combat.
Please refer to the supplementary for more details about our network architecture and training/evaluation configurations.

\subsection{Comparison to State-of-the-Arts}
\label{subsec:sota}

We compare the proposed ActFormer with the following baseline methods: Action2Motion~\cite{action2motion}, ACTOR~\cite{actor}, and CSGN~\cite{csgn}, on both single- and multi-person motion generation tasks.
Action2Motion and ACTOR can be directly evaluated on the specified datasets by simply adapting motion representations if needed.
CSGN is an unconditional motion generation method, and we extend it to conditional generation by incorporating conditional BatchNorm~\cite{condbatchnorm} in generator and projection \cite{cgan_projection} in discriminator.
All the methods above are for the single-person setting, and there is no prior works tackling the multi-person case.
Therefore, we scale these methods to the multi-person case, again by concatenating multi-person motions as a whole in each frame.

The comparison on single-person motion generation is presented in~\cref{tab:sota1p}.
Action2Motion is sub-optimal in both latent prior (frame-level) and network architecture (GRU), causing it to lag behind ours on all datasets, especially for the NTU-1P dataset.
CSGN achieves good performance on NTU-1P.
However, this method, oriented for skeleton data, suffers a performance degradation when moving from NTU-1P to SMPL-based NTURecon-1P.
ACTOR achieves excellent performance on NTU-13, but degrades significantly when faced with more challenging large-scale datasets and more strict evaluation metrics (considering root translation).
Compared to them, our method demonstrates strong adaptability to various motion representations and achieves leading performance on all datasets.
In particular, our ActFormer is the only workable method on the extremely challenging, long-tailed BABEL dataset.

As~\cref{tab:sota2p} shows, when extended to the multi-person case on NTU-2P, our advantages are more significant since no prior methods take specific designs to model human interactions.
On GTA Combat, multi-person motions are further extended to at most 5 persons, making the generation task extremely challenging.
According to the experimental results on NTU-2P, CSGN is the only baseline method showing some potential (lower FIDs) to be applied on the multi-person case.
Therefore, here we compare our method with only CSGN on the GTA Combat dataset.
With the increasing number of persons, both methods are prone to performance drops.
By contrast, our ActFormer shows a more smooth decline and outperforms CSGN on each dataset split.
Despite this, the interaction complexity in 4P/5P poses a significant challenge to our method.
Specific failure cases will be discussed in~\cref{subsec:qualitative}.

\subsection{Ablation Studies}

\begin{table}[t]
  \centering
  \begin{tabular}{l | c c c }
    \hline
    \noalign{\smallskip}
    Configuration & Acc.$\uparrow$ & FID$_{m}$$\downarrow$ & FID$_{w}$$\downarrow$ \\
    \noalign{\smallskip}
    \hline
    \noalign{\smallskip}
    (1) Gaussian latent prior & 37.55 & 17.25 & 6.01 \\
    (2) GraphConv. in Gen. & 32.97 & 18.68 & 6.09 \\
    (3) Fixed PE & 37.57 & 18.38 & 5.55 \\
    \textbf{(4) Full model} & \textbf{42.01} & \textbf{12.53} & \textbf{3.57} \\
    \noalign{\smallskip}
    \hline
  \end{tabular}
  \caption{\textbf{Ablation studes:} Several design choices on NTU-1P.}
  \label{tab:ablation_1p}
  \vspace{-0.5cm}
\end{table}

In this part, we conduct ablation experiments to study various components in our proposed framework.

\noindent{\textbf{Latent Prior.}}
Here we further verify the importance of Gaussian Process latent prior in our framework.
As in~\cref{tab:ablation_1p}~(1), replacing GP with a Gaussian latent prior leads to 4.72 FID$_m$ and 2.44 FID$_w$ increases on NTU-1P.

\noindent{\textbf{Architecture Design.}}
In the ActFormer, frame-wise human motions are encoded into vector-like token embeddings.
In~\cref{tab:ablation_1p}~(2), we evaluate another design choice: explicitly modeling skeleton topology with Graph Convolution and Graph Upsampling like in~\cite{csgn}, meanwhile modeling temporal correlations with T-Former.
It lags behind ActFormer, suggesting no need to explicitly model skeleton topology in this Transformer-based framework.

\noindent{\textbf{Positional Encoding.}}
In the data-dependent Transformer architecture, positional encoding (PE) is a crucial component which brings additional positional dependencies.
We find it better to make PE learnable rather than fixed in our method, given the results in~\cref{tab:ablation_1p}~(3) and~\cref{tab:ablation_2p}~(8).
Moreover, for multi-person case, learning a 2D combination of temporal and person-wise PE is superior to ~\cref{tab:ablation_2p}~(9) of completely learnable independent PE, showing another tradeoff between inductive bias and representation capacity.

\noindent{\textbf{Discriminator.}}
In the multi-person case, the GCN discriminator in the GAN training framework receives concatenated multi-person motions as input, equipped with a simple data augmentation strategy for permutation invariance.
Here we investigate its effectiveness by comparing with other designs which embed permutation invariance into the discriminator architecture with symmetric functions.
Specifically, these designs apply the same motion discriminator to each person independently and use the pooling/self-attention module to aggregate their features.
\cref{tab:ablation_2p}~(5-7) shows that all of them fall behind our simple combination of concatenation with data augmentation.

\begin{table}[t]
\scriptsize
  \centering
  \begin{tabular}{l | c c c c c}
    \hline
    \noalign{\smallskip}
    Configuration & Acc.$\uparrow$ & FID$_{m}$$\downarrow$ & FID$_{w}$$\downarrow$ & FID$^{a}_{m}$$\downarrow$ & FID$^{a}_{w}$$\downarrow$ \\
    \noalign{\smallskip}
    \hline
    \noalign{\smallskip}
    (5) AvgPool. in Disc. & 53.85 & 20.42 & 8.33 & 25.35 & 10.09 \\
    (6) MaxPool. in Disc. & 60.19 & 10.71 & 4.01 & 17.26 & 5.42 \\
    (7) SelfAtt. in Disc. & 51.53 & 19.96 & 8.27 & 24.91 & 8.75 \\
    (8) Fixed PE & 61.42 & 4.86 & 3.64 & 7.71 & 2.78 \\
    (9) Independent PE & 67.19 & 4.08 & 3.38 & \textbf{7.00} & 2.62 \\
    \textbf{(10) Full model} & \textbf{69.65} & \textbf{3.77} & \textbf{3.27} & 7.12 & \textbf{2.46} \\
    \noalign{\smallskip}
    \hline
  \end{tabular}
  \caption{\textbf{Ablation studies:} Several design choices on NTU-2P.}
  \label{tab:ablation_2p}
  \vspace{-0.5cm}
\end{table}

\begin{table*}[t]
  \centering
  \begin{tabular}{l | c c c c c | c c c c}
    \hline
    \noalign{\smallskip}
    & \multicolumn{5}{c}{NTU-2P} & \multicolumn{4}{c}{GTA Combat (2$\sim$5 P)} \\
    Configuration & Acc.$\uparrow$ & FID$_{m}$$\downarrow$ & FID$_{w}$$\downarrow$ & FID$^{a}_{m}$$\downarrow$ & FID$^{a}_{w}$$\downarrow$ & \multicolumn{4}{c}{FID$^a$$\downarrow$} \\
    \noalign{\smallskip}
    \hline
    \noalign{\smallskip}
    (1) w/o shared $z$ & 56.35 & 14.79 & 5.73 & 20.69 & 7.26 & 1.26 & 1.36 & 1.84 & 2.42 \\
    (2) Interaction by concatenation & 56.46 & 7.63 & 4.33 & 11.51 & 3.57 & 1.15 & 1.37 & 2.23 & 7.62 \\
    \textbf{(3) Complete model} & \textbf{69.65} & \textbf{3.77} & \textbf{3.27} & \textbf{7.12} & \textbf{2.46} & \textbf{1.04} & \textbf{1.13} & \textbf{1.58} & \textbf{2.23} \\
    \noalign{\smallskip}
    \hline
  \end{tabular}
  \vspace{-0.3cm}
  \caption{\textbf{Ablation study:} Interaction encoding on NTU-2P and GTA Combat.}
  \label{tab:ablation_interact}
\end{table*}

\noindent{\textbf{Interaction Encoding.}}
The leading performance of our method in multi-person motion generation is credited to two key designs. 
The first is to share the same latent vector sequence among multiple persons for inherent synchronization.
The other is to model human interactions with I-Former.
To verify their respective contributions, we experiment in~\cref{tab:ablation_interact}~(1) to sample different latent vector sequences for different persons independently, and~\cref{tab:ablation_interact}~(2) to adopt ActFormer for the single-person setting and output concatenated multi-person motions as a whole, without explicitly modeling interactions inside.
From experiments on NTU-2P as shown in~\cref{tab:ablation_interact}, we find both designs indispensable to synthesize high-quality interactive actions.

We further investigate the interaction encoding by experiments on GTA Combat.
Despite the performance drop with the increasing number of persons, the complete model always maintains a leading position.
When the number of persons reaches to 5, the ablation \cref{tab:ablation_interact}~(2) collapses with a 7.62 FID$_a$.
We attribute this to the fact that human interactions become sparse correlations at this moment.
Compared to position-dependent operation in concatenation-based interaction module, encoding human interactions with data-dependent self-attention in I-Former can better model such sparse correlations.

\subsection{Qualitative results}
\label{subsec:qualitative}

\begin{figure*}[t]
  \centering
   \includegraphics[width=1.0\linewidth]{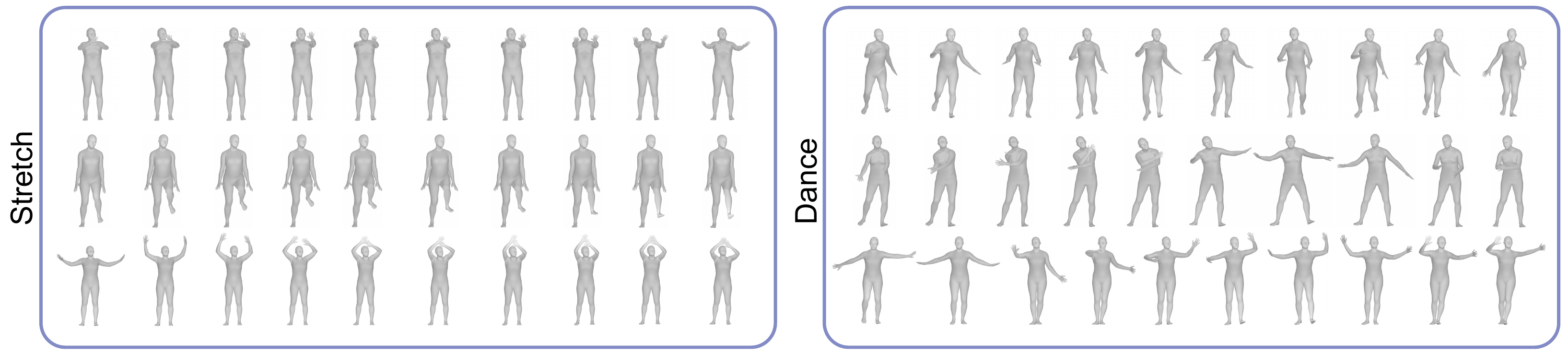}
   \caption{\textbf{Generated single-person motions.}
   The ``Stretch'' and ``Dance'' actions are both from BABEL.}
   \label{fig:gen_results_1p}
  \vspace{-0.5cm}
\end{figure*}

\begin{figure}[t]
  \centering
   \includegraphics[width=1.0\linewidth]{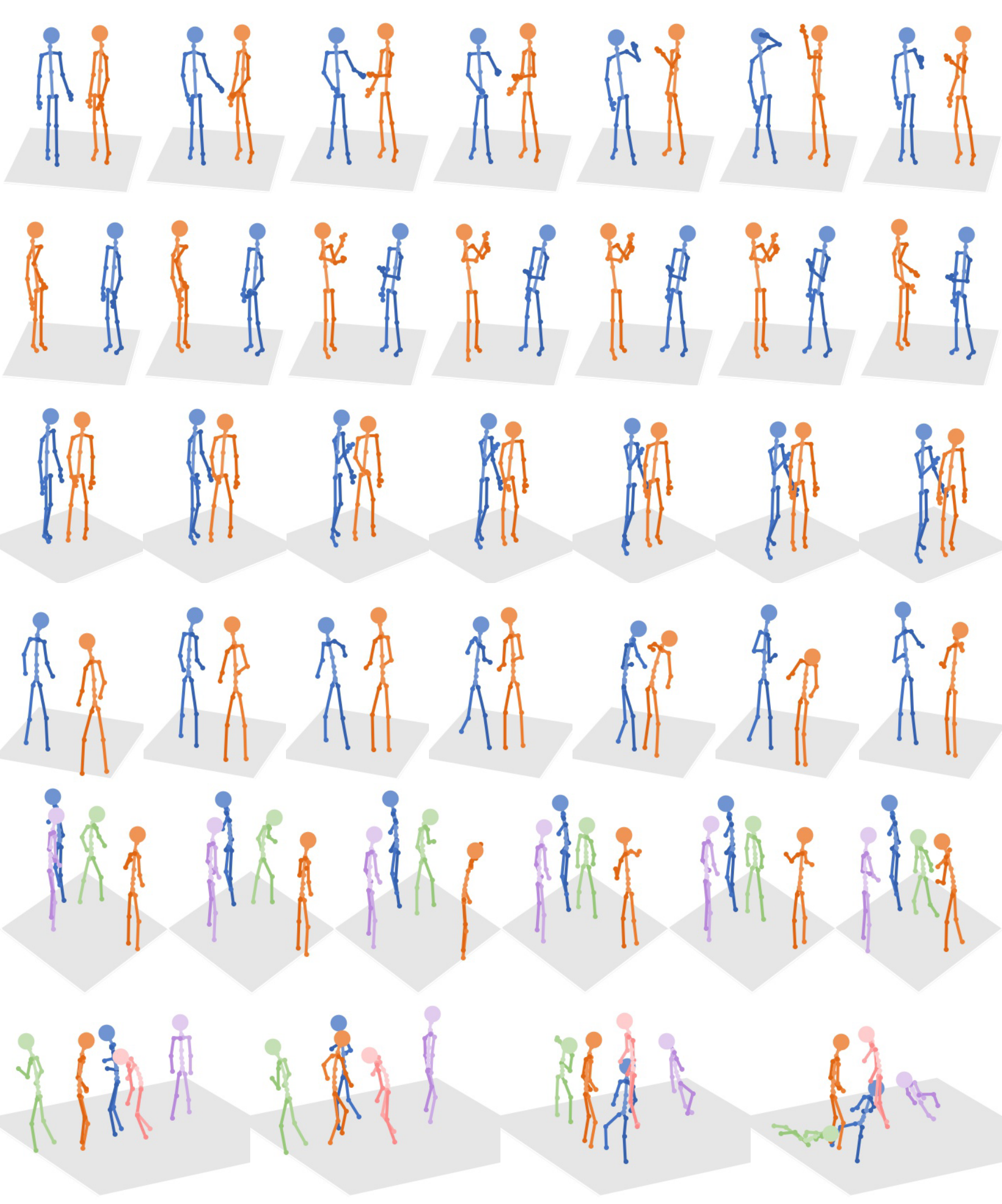}
   \caption{\textbf{Generated multi-person motions.}
   \textbf{Row 1$\sim$3:} ``Cheers and drink'', ``Take a photo'', and ``Support somebody'' actions from NTU-2P.
   \textbf{Row 4$\sim$6:} ``Combat'' actions from GTA Combat.}
   \label{fig:gen_results_2p}
  \vspace{-0.5cm}
\end{figure}

We further evaluate the generation quality of ActFormer by visualization.
Firstly, we visualize several generated samples for single-person actions from BABEL in~\cref{fig:gen_results_1p}.
For each action label, diverse motions are synthesized.
In the ``Stretch'' case, the persons are stretching different body parts in different samples.
In the ``Dance'' case, various dancing styles are presented.

In~\cref{fig:gen_results_2p} we visualize multi-person motion generation results for actions from NTU-2P and GTA Combat.
In the multi-person case, we focus more on the synchronization of human interactions.
In our generated samples, the motions of different participants are well synchronized, making the interactions look natural and vivid.
For example, in the ``Cheers and drink'' case, the two persons' toasting, cheering, and drinking actions are temporally well-matched.
In the ``Take a photo'' case, after the photographer on the left has prepared the camera, the person on the right poses up immediately (gesturing with one hand while slightly leaning back) and maintains the posture until the end of photographing.
Synchronization is reflected in not only local poses but also global trajectories.
As seen in the ``Support somebody'' case, the two persons are stumbling together.
Their trajectories keep close since the sick one relies on the other to walk.
In the first ``Combat'' case, we see the blue fighter suddenly attack the opponent and make him stagger backward.
The second ``Combat'' case is more complex, in which four persons fight in pairs.
The right two clash, while the left two, although not contacting, maintain fight postures and constantly move to find chances.
More qualitative \textbf{video results} can be found in the supplementary.

The last sample in~\cref{fig:gen_results_2p} shows a failure case.
The green and purple persons seem to interact with nobody and then fall onto the ground without being attacked.
This case reflects a limitation of our method: when the number of persons increases, human interactions become sparse and may form two separate groups.
ActFormer has no mechanism to divide persons into different interaction groups, thus not sufficiently effective to learn from some GTA Combat data with 4 or 5 persons.

\section{Conclusion and Discussion}

This work explores a solution towards general action-conditioned human motion generation and proposes ActFormer, a GAN-based Transformer framework.
The ActFormer is evaluated on several challenging benchmarks and achieves leading performance over prior methods on various human motion representations and both single-person and multi-person motion generation tasks.
Detailed ablation studies are also conducted to investigate the various components in our approach.
The ActFormer adapts to a more complete domain of human actions compared to prior works, while the general human motion generator is still not reached.
Human-object interaction synthesis remains unexplored, and we leave this direction for future exploration.


\noindent{\textbf{Broader impacts.}} The proposed generative method can synthesize non-existing content. The community should be wary of the malicious uses of this feature. The collected GTA Combat dataset contains fighting scenes, while we do not promote violence. The dataset should only be used for research on modeling multi-person interactive actions.






\clearpage
\appendix

\section*{Appendices}

We provide the details about our network architecture (\cref{sec:net_arch}), additional implementation details (\cref{sec:more_implement}), additional qualitative results (\cref{sec:more_qual}), additional samples from our GTA Combat dataset (\cref{sec:more_gta}) and the limitations of our method (\cref{sec:limitations}).



\section{Network architecture}
\label{sec:net_arch}

\begin{table*}[t]
  \centering
  \begin{tabular}{l l l}
    \hline
    \noalign{\smallskip}
    Network & Architecture & Params. \\
    \noalign{\smallskip}
    \hline
    \noalign{\smallskip}
    \multirow{7}{*}{Generator} & (Input projection): Linear(Ch=(120, 200)) & \multirow{7}{*}{1,359,675} \\
    & (Class embedding): Linear(Ch=(26, 200)) & \\
    & (I-Former): Transformer(Ch=200, n\_heads=8) & \\
    & (T-Former): Transformer(Ch=200, n\_heads=8) & \\
    & (I-Former): Transformer(Ch=200, n\_heads=8) & \\
    & (T-Former): Transformer(Ch=200, n\_heads=8) & \\
    & (Output projection): Linear(Ch=(200, 75)) & \\
    \noalign{\smallskip}
    \hline
    \noalign{\smallskip}
    \multirow{7}{*}{Discriminator} & GraphConv.(Ch=(150, 32), K=(2, 4), G=(25, 25)) & \multirow{7}{*}{5,174,593} \\
    & GraphConv.(Ch=(32, 64), K=(2, 4), G=(25, 11)) & \\
    & GraphConv.(Ch=(64, 128), K=(2, 4), G=(11, 5)) & \\
    & GraphConv.(Ch=(128, 256), K=(2, 4), G=(5, 5)) & \\
    & GraphConv.(Ch=(256, 512), K=(5, 4), G=(5, 1)) & \\
    & (Class embedding): Linear(in\_ch=26, out\_ch=512) & \\
    & (Output projection): Linear(in\_ch=512, out\_ch=1) & \\
    \noalign{\smallskip}
    \hline
  \end{tabular}
  \caption{\textbf{Network architecture on NTU-2P dataset.}
  Params. is short for the \emph{number of parameters}.
  The tuple for \emph{Ch (channel)} denotes input/output channels of the layer.
  In GraphConv., the tuple for \emph{K (kernel)} denotes spatial/temporal kernel sizes of the graph convolution layer.
  The tuple for \emph{G (graph)} represents the number of nodes (joints) in the input/output spatial skeleton graph.
  }
  \label{tab:net_arch}
\end{table*}

\cref{tab:net_arch} presents the network architecture we designed on the NTU-2P dataset.
In the ActFormer generator, Gaussian Linear Error Units (GELU) \cite{gelu} activations are used for Transformer encoders.
In the GCN discriminator, Leaky Rectified Linear Units (LeakyReLU) \cite{leakyrelu} activations are adopted.
This network architecture can also be applied to other datasets.
The dimension of inputs/outputs may slightly vary due to changes in pose representations or the number of persons.
On single-person action datasets, I-Former modules in the ActFormer generator are not needed.

\paragraph{I-Former and T-Former.}

Here we give a detailed illustration of the I-Former and T-Former modules.
In both modules, given $P \cdot (T+1)$ tokens (corresponding to a T-frame, P-person motion sequence) as input, we represent the current token embedding by $F_{t}^{p} \in \mathbb{R}^{C_{in}}$, where $t \in [1, ..., T+1]$ and $p \in [1, ..., P]$.
First, we apply linear transformations to each token to compute \{query, key, value\} vectors $q_{t}^{p}, k_{t}^{p}, v_{t}^{p} \in \mathbb{R}^{d}$ as,
\begin{equation}
    q_{t}^{p} = W_q F_{t}^{p}, k_{t}^{p} = W_k F_{t}^{p}, v_{t}^{p} = W_v F_{t}^{p},
\end{equation}
where the learnable transform parameters $W_q, W_k, W_v \in \mathbb{R}^{d \times C_{in}}$ are shared among all tokens.
Then we discuss subsequent steps in I-Former and T-Former, respectively.

In the I-Former module, we apply self-attention among persons in every single frame independently by query-key dot product and the weighted sum of values as below,
\begin{equation}
    w_{t}^{p,p'} = q_{t}^{p} \cdot k_{t}^{p'}, \forall t \in [1, ..., T+1],
\end{equation}
\begin{equation}
    G_{t}^{p} = \sum_{p'} softmax_{p'} (\frac{w_{t}^{p,p'}}{\sqrt{d}})v_{t}^{p'}.
\end{equation}
Now we got the transformed token embedding $G_{t}^{p} \in \mathbb{R}^{d}$ by the I-Former.

In the T-Former, self-attention is performed among frames of each person separately.
To be specific, the T-Former transforms the token embedding into $H_{t}^{p} \in \mathbb{R}^{d}$ as in the following,
\begin{equation}
    w_{t, t'}^{p} = q_{t}^{p} \cdot k_{t'}^{p}, \forall p \in [1, ..., P],
\end{equation}
\begin{equation}
    H_{t}^{p} = \sum_{t'} softmax_{t'} (\frac{w_{t, t'}^{p}}{\sqrt{d}})v_{t'}^{p}.
\end{equation}

\begin{figure}[t]
  \centering
  \includegraphics[width=0.8\linewidth]{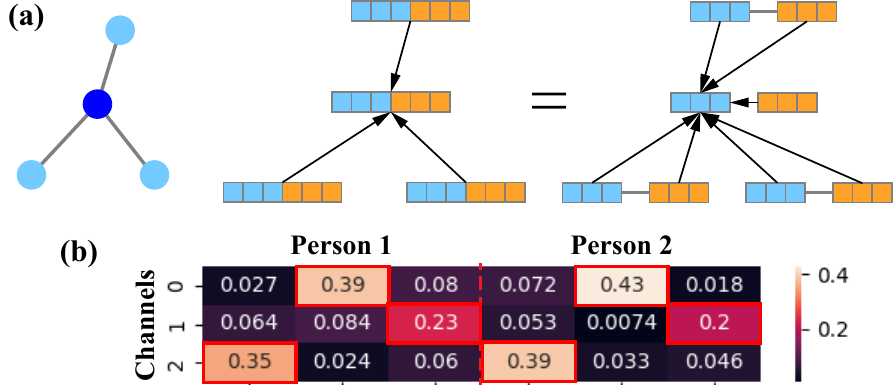}
  \caption{\textbf{Illustration about interaction modeling in GCN discriminator.}
  We make a theoretical analysis about (a) how concatenation enforces information aggregation from multiple persons. 
  We also check (b) the weights of a trained discriminator to verify whether information from multiple persons is aggregated in practice.}
  \label{fig:gcn_cat}
\end{figure}

\paragraph{Interaction modeling in GCN discriminator.}
Here we illustrate in detail how concatenation models multi-person interactions in our GCN discriminator.
We drop the temporal dimension $T$ for simplicity and observe a $K$-node graph of the spatial human skeleton.
Each node contains a vector $f^k \in \mathbb{R}^{P \cdot D}$ concatenated by $P$ persons.
We focus on a specific node with $K_N$ neighbors in the skeleton, as shown in~\cref{fig:gcn_cat}(a) ($K_N=4$ including the center node itself, $P = 2, D = 3$ in this example).
A spatial GraphConv. kernel $w$ aggregates information from all the neighboring nodes and produce an output feature $g$ for the center node, i.e.,
\begin{equation}
  g = \sum_{k=1}^{K_N} \sum_{l=1}^{P \cdot D} w^k_l \cdot f^k_l.
  \label{eq:gcn_cat}
\end{equation}
If we split multiple persons' motion in the same node, i.e., transform  $f^k \in \mathbb{R}^{P \cdot D}$ into  $f^k \in \mathbb{R}^{P \times D}$, \cref{eq:gcn_cat} can be rewritten as
\begin{equation}
  g = \sum_{k=1}^{K_N} \sum_{l=1}^{P \cdot D} w^k_l \cdot f^k_l = \sum_{k=1}^{K_N} \boldsymbol{\sum_{p=1}^{P}} \sum_{d=1}^{D} w^k_{pd} \cdot f^k_{pd}.
\end{equation}
As visualized in~\cref{fig:gcn_cat}(a), the output center node feature $g$ aggregates information from all the $K_N$ neighboring joints of all the $P$ persons.

Besides theoretical analysis, we also check absolute values of trained weights from our GCN discriminator’s 1st layer.
As shown in~\cref{fig:gcn_cat}(b), each output node feature indeed gets contributions from multiple persons.
The information of different persons has been entangled since then, and human interaction modeling naturally exists in the following network layers.

\section{Additional implementation details}
\label{sec:more_implement}

\paragraph{Pose representations.}
On NTU RGB+D 120 and GTA Combat, local body poses are represented by normalized limb vectors instead of raw joint coordinates.
On NTU-13 and BABEL, the continuous 6D representations \cite{6d_rot} are employed to replace axis-angle joint rotations from SMPL body models.

\paragraph{Library credits.}
Our approach is implemented based on PyTorch \cite{pytorch}.
For comparison to baseline methods, we use the official implementations of Action2Motion~\cite{action2motion_code} and ACTOR \cite{actor_code}.

\paragraph{Training.}
In the ActFormer's training, we adopt the Adam optimizer with betas (0, 0.999) and learning rate 0.0002 for both the generator and discriminator.
The batch size is set to 64.
During training, every time the discriminator is trained 4 times, the generator will be trained once.
On each dataset, we train all models (including baseline methods) for 500 epochs.

In the training of the action recognition model, we adopt the SGD optimizer,  with an initial learning rate of 0.1 decayed by 0.1 in the 10th and 50th epoch, respectively.
The model is trained for 80 epochs on each dataset, and the batch size is set to 64.

\paragraph{Evaluation.}
We use each model to generate a sample set with 100 sequences per action category for evaluation.
All motion sequences in our experiments hold 60 frames.

\section{Additional qualitative results}
\label{sec:more_qual}

\begin{figure*}[t]
  \centering
  \includegraphics[width=1.0\linewidth]{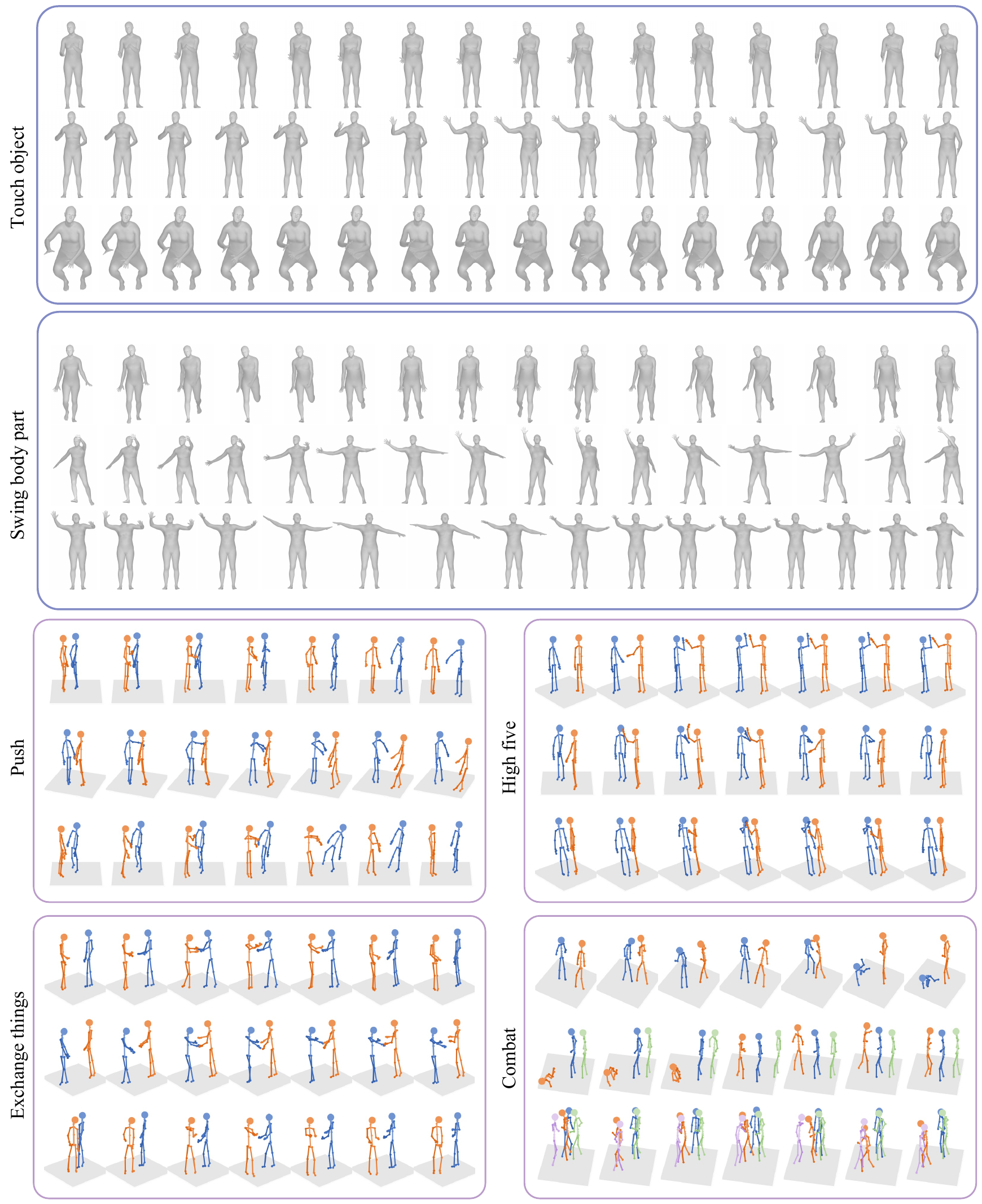}
  \caption{\textbf{Additional qualitative results.}
  We generate "Touch object" and "Swing body part" actions from BABEL, "Push" "High five" and "Exchange things" actions from NTU-2P, and "Combat" actions from GTA Combat.}
  \label{fig:gen_results_supple}
\end{figure*}

\cref{fig:gen_results_supple} displays more generated motions by our approach.
Compared to qualitative results in the main paper, we show more samples per action for multi-person actions to indicate that diversity also exists in our multi-person motion generation.

\section{Additional GTA Combat samples}
\label{sec:more_gta}

\begin{figure*}[t]
  \centering
  \includegraphics[width=1.0\linewidth]{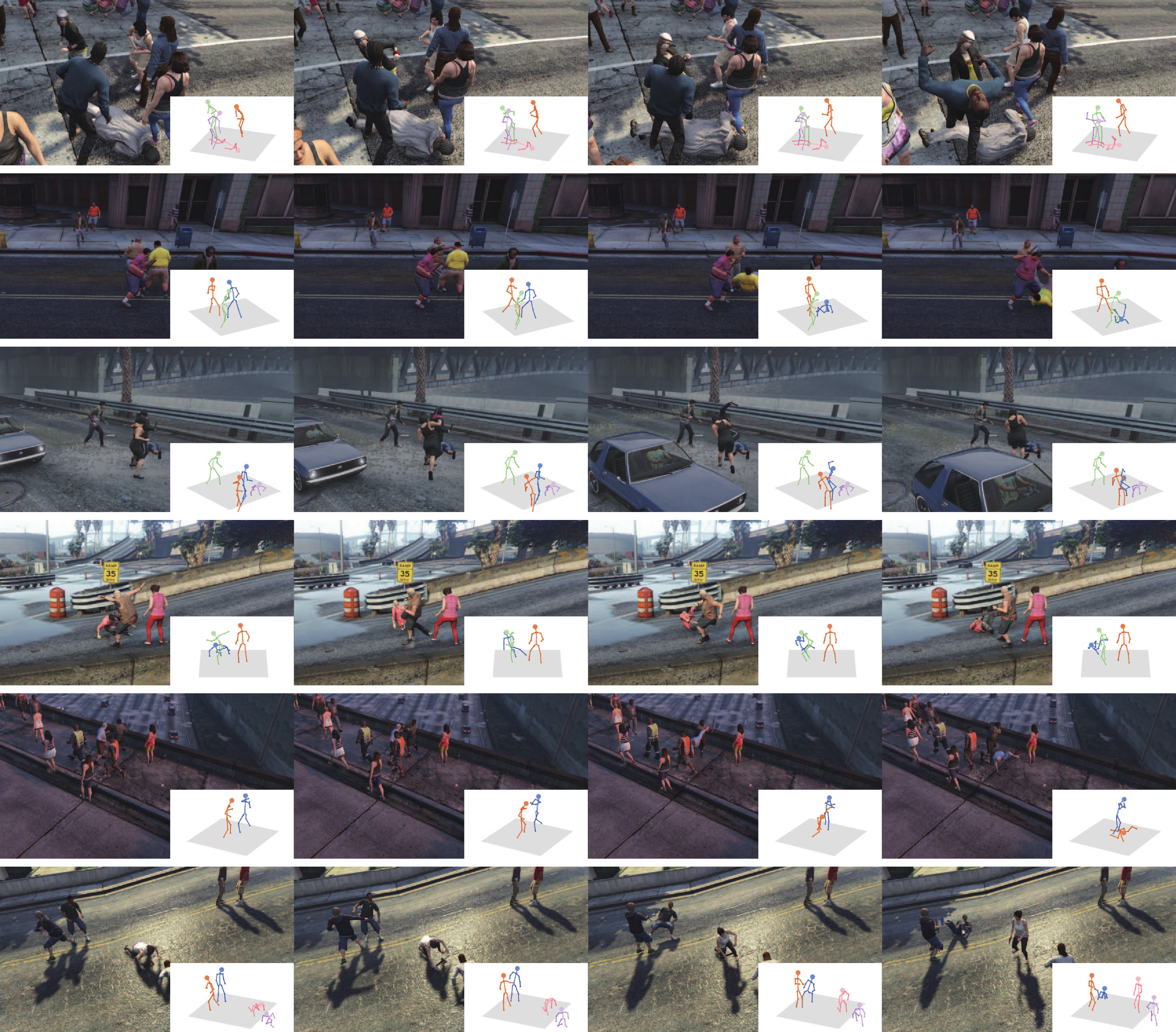}
  \caption{\textbf{Additional samples from GTA Combat dataset.}}
  \label{fig:gta_combat_supple}
\end{figure*}

We provide more samples from our GTA Combat dataset.
As shown in \cref{fig:gta_combat_supple}, by randomly combining interactive relationships and combat actions, the synthetic motions in GTA Combat present sufficient diversity.
From dynamic clips, we see both combat actions of attackers and triggered reactions of attacked ones look natural.

\section{Limitations}
\label{sec:limitations}
As discussed in the qualitative results section of the main manuscript, our method has no mechanism to divide persons into different interaction groups. Therefore, given MoCap samples in which multiple persons form separate interaction groups, the ActFormer cannot effectively learn from it. Besides, in the GAN training, multi-person motions are concatenated before input to the GCN discriminator. Thus we cannot learn a shared model for motions with a variable number of persons, despite the ActFormer being a Transformer-based generator.

\end{document}